\pdfoutput=1

\documentclass[11pt]{article}

\usepackage{EMNLP2023}

\usepackage{times}
\usepackage{latexsym}
\usepackage{booktabs}
\usepackage{multirow}
\usepackage{ulem}
\usepackage{graphicx}
\usepackage{booktabs}
\usepackage{multirow}
\usepackage{amsmath}
\usepackage{float}
\usepackage{algorithm}
\usepackage{algpseudocode}
\usepackage{amsmath}
\usepackage{amsfonts}
\usepackage{multicol}
\usepackage{lipsum}
\usepackage{tabto}

\usepackage[T1]{fontenc}

\usepackage[utf8]{inputenc}

\usepackage{microtype}

\usepackage{inconsolata}

\title{SeRTS: Self-Rewarding Tree Search for Biomedical \\ Retrieval-Augmented Generation}

\author{
  Minda Hu$^{1}\thanks{\ \ The first two authors have equal contributions.}$\hspace{0.3cm}Licheng Zong$^{1}$\footnotemark[1]\hspace{0.3cm}Hongru Wang$^2$\hspace{0.3cm}Jingyan Zhou$^2$\hspace{0.3cm}Jingjing Li$^1$\\
  \bf
  Yichen Gao$^1$\hspace{0.3cm}Kam-Fai Wong$^2$\hspace{0.3cm}Yu Li$^1$\hspace{0.3cm}Irwin King$^1$\\
  $^1$Department of Computer Science and Engineering\\
  $^2$Department of Systems Engineering and Engineering Management\\
  The Chinese University of Hong Kong\\
  \texttt{\{mindahu21, lczong21, liyu, king\}@cse.cuhk.edu.hk}
  }

\begin{document}
\maketitle
\begin{abstract}
Large Language Models (LLMs) have shown great potential in the biomedical domain with the advancement of retrieval-augmented generation (RAG). However, existing retrieval-augmented approaches face challenges in addressing diverse queries and documents, particularly for medical knowledge queries, resulting in sub-optimal performance. To address these limitations, we propose a novel plug-and-play LLM-based retrieval method called \textbf{Se}lf-\textbf{R}ewarding \textbf{T}ree \textbf{S}earch (\textbf{SeRTS}) based on Monte Carlo Tree Search (MCTS) and a self-rewarding paradigm. By combining the reasoning capabilities of LLMs with the effectiveness of tree search, SeRTS boosts the zero-shot performance of retrieving high-quality and informative results for RAG. We further enhance retrieval performance by fine-tuning LLMs with Proximal Policy Optimization (PPO) objectives using the trajectories collected by SeRTS as feedback. Controlled experiments using the BioASQ-QA dataset with GPT-3.5-Turbo and LLama2-7b demonstrate that our method significantly improves the performance of the BM25 retriever and surpasses the strong baseline of self-reflection in both efficiency and scalability. Moreover, SeRTS generates higher-quality feedback for PPO training than self-reflection. Our proposed method effectively adapts LLMs to document retrieval tasks, enhancing their ability to retrieve highly relevant documents for RAG in the context of medical knowledge queries. This work presents a significant step forward in leveraging LLMs for accurate and comprehensive biomedical question answering.

\end{abstract}

\section{Introduction}
Large Language Models (LLMs) have made significant advancements and gained widespread adoption in various natural language tasks, such as dialogue systems \cite{bang-etal-2023-multitask, wang-etal-2023-large} and question-answering \cite{kamalloo2023evaluating, li-etal-2023-interview,selfdc}. However, their performance in specialized domains like medicine remains unsatisfactory due to the lack of publicly available training data, which limits the amount of domain-specific knowledge these LLMs can encode in their parameters \cite{xia-etal-2022-medconqa, shi2024medical}. To address this limitation, recent studies have turned to retrieval-augmented generation (RAG) frameworks \cite{xiong2024benchmarking, unims_rag, gao2024enhancing}. These frameworks can enhance LLMs' capabilities by retrieving relevant documents from large and private databases, thereby enabling the models to provide more accurate and informative answers.

Existing RAG framework mainly consists of two modules: an external retriever to retrieve documents according to semantic similarity between documents and queries, and the reader to generate the final answer given retrieved documents and query \cite{gao2024retrievalaugmented}. However, this kind of setting overlook the two significant challenges in medical domain: the complexity of the queries and the varying importance of different documents. For example, patients may describe the same disease in diverse ways, leading to variations in query formulation. Additionally, documents from different medical disciplines should be weighted differently based on both of their relevance and importance. Thus, effectively identifying the most pertinent and useful documents from a vast corpus of medical literature is crucial. This not only enhances performance but also improves the explainability and credibility of the system.

To address these issues, we propose a plug-and-play LLM-based retrieval method, \textsc{Self-Rewarding Tree Search~(SeRTS)}, combines the exceptional reasoning capabilities of LLMs with the effective searching capabilities of Monte Carlo Tree Search (MCTS). Specifically, we first formulate the task of document retrieval as a tree search problem, considering both the diversity of queries and different relationship between multiple documents. Inspired by recent studies using MCTS to search the optimal reasoning paths of LLMs, we alternativelly employ it to find the best query and most related documents from vast candidate spaces. To ensure the success of MCTS, we carefully design each operations in MCTS and provide detailed feedback from different levels of documents. Furthermore, to validate the effectivenss and quality of our tree search retriever, we then collect these feedbacks with all intermediate reasoning results to finetune LLMs using PPO method. To sum up, our contributions can be summarized as follows:

\begin{itemize}
    \item We present a novel perspective on automating query optimization by combining LLM and MCTS for improved document retrieval in the medical domain.
    \item We propose a Self-Rewarding Tree Search framework, consisting of a tree search retriever and a PPO self-training process.
    \item Experimental results on two LLMs, \textsc{GPT-3.5-Turbo} and \textsc{LLama2-7b-chat}, demonstrate that our SeRTS method significantly improves the performance of BM25 retrievers and surpasses the strong baseline of self-reflection in both efficiency and scalability.
\end{itemize}

\section{Related Works}
\paragraph{Medical RAG.} Retrieval-Augmented Generation represents a significant advancement to provide up-to-date and trustworthy collections and improves the transparency of LLMs by grounding their reasoning on the retrieved documents \citep{xiong2024benchmarking}, particularly in the medical question-answering \cite{tian-etal-2019-chimed, xia-etal-2022-medconqa} and dialogue tasks \cite{shi2024medical}. Most of previous works focus on query rewriting \cite{zhang-etal-2022-focus-driven} and multi-step retrieval \cite{mrini-etal-2022-medical}. More recent studies explore how LLMs can improve literature information-seeking and clinical decision-making RAG  \cite{zakka2023almanac, wang2024augmenting, jeong2024improving}. In detail, \citet{wang2024augmenting} design a hybrid retriever and conduct complex filter operations to get high-quality documents while \citet{jeong2024improving} propose Self-BioRAG to utilize retrieval and self-reflection together, resulting in better medical reasoning. 

\paragraph{Applications of MCTS.} Monte Carlo Tree Search (MCTS) has been validated as an effective method for finding better solution paths among a wide range of potentially feasible options, enabling proficient exploration and enhancing decision-making\citep{silver2017mastering}.  It has proven to be highly successful in various domains, including game-playing~\cite{sironi2018self,ontanon2016informed}, and planning~\cite{zhou2023language,yu-etal-2023-prompt}. By combining MCTS with reinforcement learning, models can learn from self-play and achieve human-level performance or even surpass it in complex tasks, such as the game of Go~\cite{silver2016mastering}. Recent works mainly utilize MCTS to determine the most promising reasoning path of LLMs, leading to better performance without additional annotations \cite{feng2024alphazerolike, zhou2024language, tian2024selfimprovement}. Specifically, \citet{feng2024alphazerolike} fix a search step to be either a token or a sentence while \citet{tian2024selfimprovement} employ $\eta$MCTS to search across multiple levels of planning with carefully defined reward signals. Additionally, other studies use MCTS to collect high-quality reasoning traces to train the reward model, aiming to continuously enhance LLMs over multiple iterations \citep{zhang2024restmcts}. Furthermore, \citet{zhou2024language} expands the search space to encompass all possible reasoning and acting steps, thereby unifying reasoning, acting, and planning in language models.

\paragraph{Learning from Feedback.} LLMs can evolve and then refine its outputs by learning from self-assessed rewards or environmental signals, such as self-learning or reinforcement learning from human feedback (RLHF). On the one hand, self-learning methods leverage the LM's ability to critique its outputs, collecting natural language feedback as signals, such as Reflection \cite{shinn2024reflexion}, Self-Rewarding LMs \cite{yuan2024selfrewarding}. On the other hand, lots of works apply reinforcement learning to train the external reward model from human preferences \cite{ouyang2022training}. The reward model is then frozen and used to train the LLM using RL, e.g., Proximal Policy Optimization (PPO) \cite{schulman2017proximal, wang2022integrating}, which is a popular reinforcement learning algorithm that has been successfully applied to various domains, including natural language processing tasks~\cite{han-etal-2023-dialcot,akyurek-etal-2023-rl4f}. However, most existing reinforced self-improvement approaches, e.g., STaR \cite{zelikman2022star} and RFT \cite{yuan2023scaling}, overlooking the possibility that incorrect or low-quality reasoning traces may still lead to the correct solution due to hallucinations or randomness, leading to an unreliable reward model \cite{lanham2023measuring}.

\section{Method}
\begin{figure}[h!]
    \centering
    \includegraphics[width=0.45\textwidth]{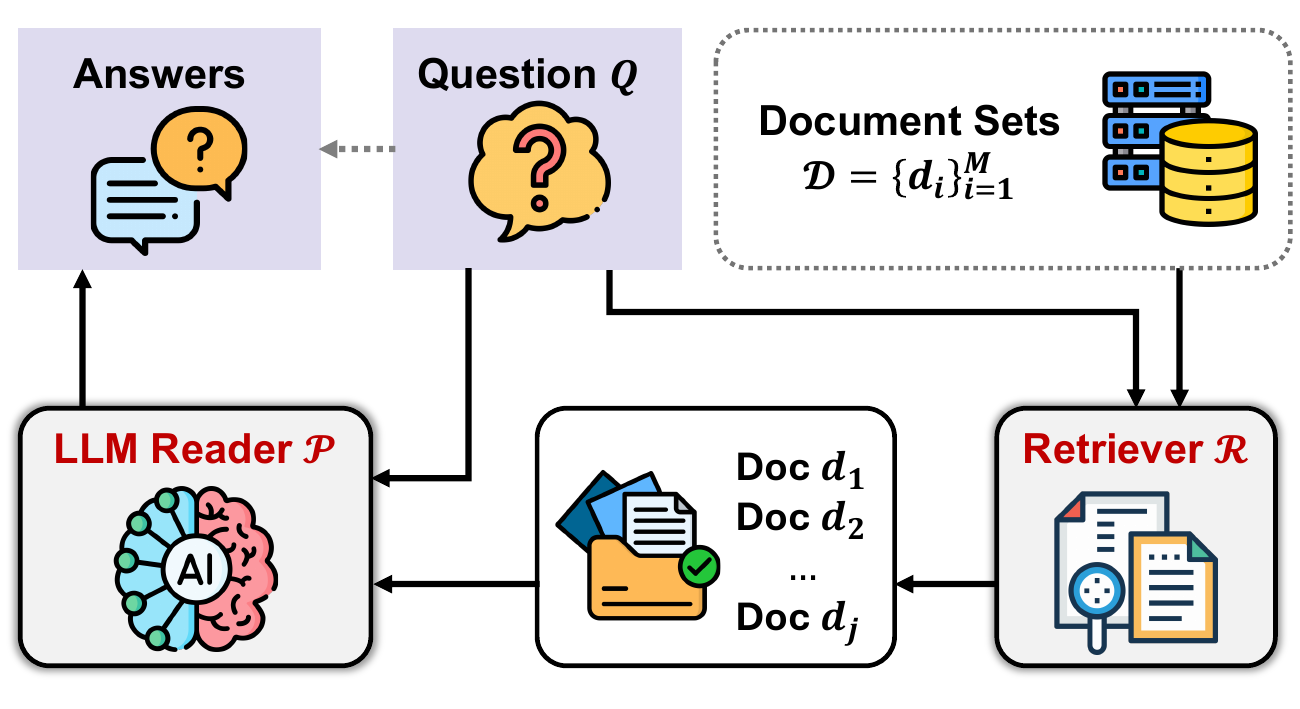}
    \caption{A straightforward Retrieval-Augmented Generation (RAG) pipeline}
    \label{fig:rag}
\end{figure}

\begin{figure*}[t!]
    \centering
    \includegraphics[width=\textwidth]{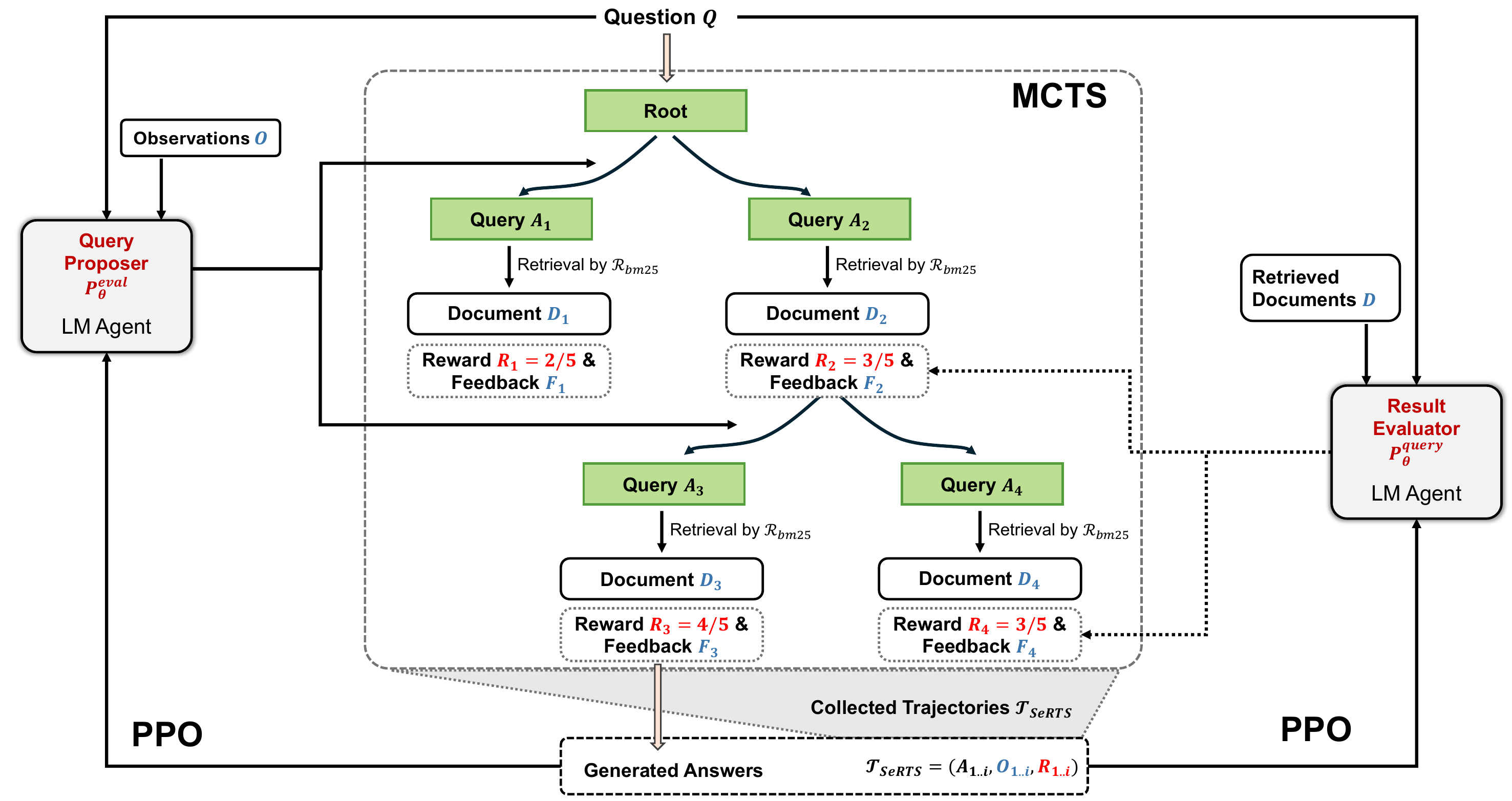}
    \caption{SeRTS method overview: \textbf{Query Proposer} $P^{query}_{\theta}$ generates query $A_i$. $\mathcal{R}_{bm25}$ retrieves relevant documents $D_i$. \textbf{Result Evaluator} $P^{eval}_{\theta}$ assesses $D_i$, provides Reward $R_i$ and Feedback $F_i$. Observations $O_i$ (previous queries, retrieved documents, feedback) serve as input to $P^{query}_{\theta}$ for subsequent query proposals. The entire trajectory $\mathcal{T}_{SeRTS}$ (initial question $Q$, $A_{1..i}$, $O_{1..i}$, $R_{1..i}$) is used to fine-tune the two language model agents via PPO, to improve their performance in query proposal and evaluation.}
    \label{fig:overview}
\end{figure*}

Figure \ref{fig:rag} illustrates a straightforward Retrieval-Augmented Generation (RAG) pipeline we employed as the basis of our experiments. This process can be defined as follows:
Give a literature document set $\mathcal{D}=\{d_i\}_{i=1}^M$, a Language Model based reader $\mathcal{P}$, an external retriever $\mathcal{R}$, and a question $Q$, use $\mathcal{R}$ to retrieve several documents $\{d_{1..j}\}$ related to $Q$, so that the reader $\mathcal{P}$ can generate appropriate answers using them. When  $\mathcal{P}$ is fixed, the quality of the generated answers mainly depends on the retrieved $\{d_{1..j}\}$. 

To improve the relevance of $\{d_{1..j}\}$ to $Q$ for higher quality of the answers, we propose a novel method for retrieval, called \textbf{Se}lf-\textbf{R}ewarding \textbf{T}ree \textbf{S}earch (\textbf{SeRTS}) as shown in Figure~\ref{fig:overview} illustrating the Expansion \& Evaluation process in the MCTS and its interaction with two LLM agents. The core of SeRTS is a search algorithm based on Monte Carlo Tree Search (MCTS) that learns from trial-and-error experiences based on Self-Rewarding Paradigms and explores better trajectories of self-reflection reasoning to retrieve more relevant documents for Retrieval Augmented Generation (RAG). 

It is worth noting that our method is decoupled from the RAG pipeline. All components in the pipeline, including the document set $\mathcal{D}=\{d_i\}_{i=1}^M$, the LM reader $\mathcal{P}$ and the retriever $\mathcal{R}$ could be replaced by other types. Our method acts as a plugger that can be integrated into any RAG pipelines. 

\subsection{Self-Rewarding Tree Search}
As a decision policy, MCTS effectively balances exploration and exploitation when searching for optimal trajectories, so it's employed to find the best retrieval results in our proposed method.

We formulate the document retrieval process as a tree search problem. In the tree shown in Figure~\ref{fig:overview}, each node $s_i = [Q, A_{i}, D_i, R_i, F_i, O_{i}]$ represents a state with the original question $Q$, query proposal $A_{i}$, retrieved documents $F$, result score $R$ and feedback $F$, and previous process observation $O_{i}$ 
(query proposals from all $s_i$'s ancestor nodes $O^{anc}_q = \bigcup_{k \in \text{anc}(s_i)}A_k$,
retrieved documents from all $s_i$'s ancestor nodes $O^{anc}_d = \bigcup_{k \in \text{anc}(s_i)}D_k$,
query proposals from all $s_i$'s sibling nodes $O^{sib}_q = \bigcup_{k \in \text{sib}(s_i)}A_k$,
feedback from all $s_i$'s sibling nodes $O^{sib}_f = \bigcup_{k \in \text{sib}(s_i)}F_k$).

Starting from the root node (initial state $s_0$), SeRTS uses the MCTS algorithm that iteratively searches through the Query Space $A_{1..i}$ and uses $A_{1..i}$ to retrieve documents. The retriever employed here is the BM25, a strong retriever in the biomedical retrieval domain, supported by multiple evaluations~\cite{xiong2024benchmarking}. The final target is to find node $s_o$ that has retrieved documents from itself and its ancestors are evaluated to have the highest scores in helping generate answers for question $Q$.

\subsubsection{Result Evaluator for Quantitatively Measuring}
Quantitatively measuring the quality of retrieved documents by providing high-quality feedback with a specific score is essential for the MCTS. Here, we propose to utilize LMs to automatically do the evaluation and provide feedback, inspired by the Self-Rewarding paradigm~\cite{yuan2024selfrewarding}. The quality of feedback plays a crucial role in producing accurate reasoning, and low-quality feedback can hinder the self-improving ability of LLMs during self-reflection.

We design a \textbf{Result Evaluator} $P_\theta^{eval}$ using the system prompt in Table~\ref{tab:eval_prompt} to provide descriptional feedback and quantitative measurement for retrieved documents.

It utilizes the 5-point rubrics to assess the relevance of retrieved documents to the query and their potential contribution to the reasoning process, as well as providing descriptional feedback to the search algorithm. This scoring mechanism enables MCTS to make informed decisions during the search process, guiding it toward retrieving the most informative and relevant documents for RAG.

\begin{table}[!ht]
\small
    \centering
    \colorbox{orange!8}{
    \begin{tabular}{@{}p{7.2cm}}
    You are a medical expert. Review the user's question and rate the corresponding retrieved knowledge using the additive 5-point scoring system described below. Points are accumulated based on the satisfaction of each criterion:\\
    \textbf{\{Five-point Rubrics\}}\\
    \\
    <question> \textbf{\{$Q$\}} </question>\\
    <retrieved\_knowledge>\\
    \textbf{\{$D \cup O^{anc}_d$\}}\\
    </retrieved\_knowledge>\\
    \\
    After examining the user’s instruction and the response:\\
    -Briefly justify your total score, up to 100 words. Remember to assess from the AI Assistant perspective, utilizing web search knowledge as necessary. To evaluate the response in alignment with this additive scoring model, we’ll systematically attribute points based on the outlined criteria. \\
    -Give suggestions for constructing better queries up to 100 words.\\
    \\
    Conclude with a score between 0 and 5, strictly using the aforementioned additive 5-point scoring system and the format: “<score> Integer Score </score>”. For example, <score>3</score>.
    \end{tabular}}
    \caption{Prompt for Evaluator $P^{eval}_{\theta}$. Details of \textbf{\{Five-point Rubrics\}} is shown in Appendix~\ref{sec:rubric}.}
    \label{tab:eval_prompt}
\end{table}


\subsubsection{Query Proposer for Efficiently Searching}
The vast number of potential token combinations in natural language tasks results in an exponentially large search space, posing a significant challenge to the efficiency of MCTS. Our approach tackles the search space issue by employing strategic expanding techniques and heuristics to navigate the vast query space $A_{1..i}$ efficiently. 

We design a \textbf{Query Proposer} $P_\theta^{query}$ using the system prompt in Table~\ref{tab:query_prompt}. Query Proposer efficiently explores new queries in a heuristic manner using previous results and feedback. At every node, $P_\theta^{query}$ is given the query proposals from its sibling nodes and the corresponding feedback produced by the Result Evaluator $P_\theta^{eval}$, as well as previously retrieved documents from all its ancestor nodes. Considering the known information, $P_\theta^{query}$ is prompted to generate new query proposals efficiently for the next step.

\begin{table}[!ht]
\small
    \centering
    \colorbox{orange!8}{
    \begin{tabular}{@{}p{7.2cm}}
    You are a medical expert, and you are tasked with searching for as much useful knowledge from a database of medical paper abstracts as possible to address the user’s question. The question is stated as follows: \textbf{\{$Q$\}}\\
    \\
    You are given the BM25 retriever as the search tool. According to previous query proposals and proposals' corresponding feedback and suggestions, give one single better query for the retriever to further search for extra information needed for answering the given question.\\
    \\
    <query\_proposal>\\
    1. \textbf{\{$O^{sib}_{q, 1}$\}}\ \ \ \ Corresponding Feedback: \textbf{\{$O^{sib}_{f, 1}$\}}\\
    \ \ \ \ ......\\
    i. \textbf{\{$O^{sib}_{q, i}$\}}\ \ \ \ \ Corresponding Feedback: \textbf{\{$O^{sib}_{f, i}$\}}\\
    </query\_proposal>\\
    \\
    Information already known is as follows:<retrieved\_info>\\
    \ \ \ \ Query and its reasoning for Abstracts: \textbf{\{$O^{anc}_q$\}}\\
    \ \ \ \ Retrieved Abstracts: \textbf{\{$O^{anc}_d$\}}\\
    
    </retrieved\_info>\\ 
    \\
    After examining the information above, justify your decision, in up to 100 words.\\
    \\
    Hints: \\
    1. Use simple keywords as queries if no information is retrieved.\\
    2. You can take one step at a time, subdivide the original question, and search for information on the sub-questions.\\
    \\
    Conclude with the query for the paper abstracts using the format: "<query> Query Here </query>".
    \end{tabular}}
    \caption{Prompt for Query Proposer $P^{query}_{\theta}$.}
    \label{tab:query_prompt}
\end{table}


\subsubsection{Monte Carlo Tree Search Process}
\begin{figure*}[t!]
    \centering
    \includegraphics[width=\textwidth]{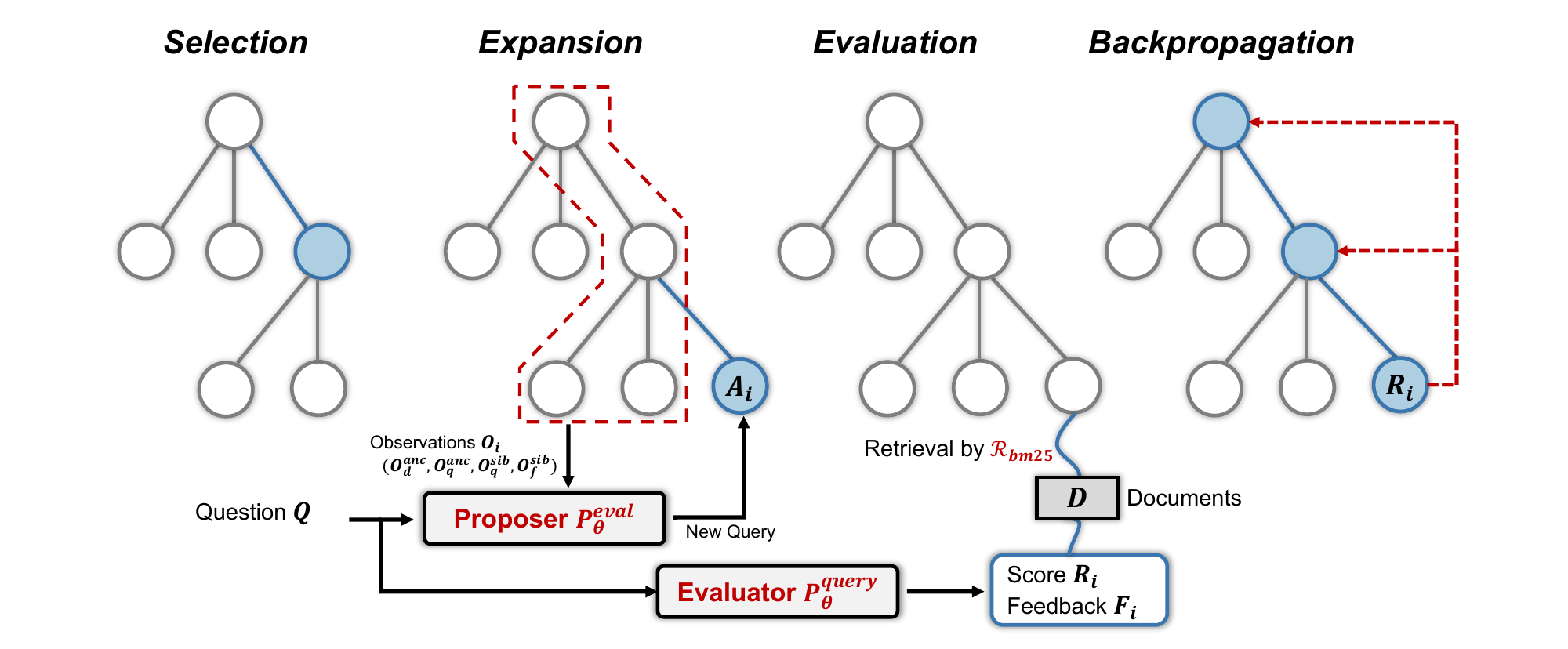}
    \caption{Overview of the four operations in MCTS}
    \label{fig:mcts}
\end{figure*}

As shown in Figure~\ref{fig:mcts}, using the \textbf{Query Proposer} $P_\theta^{query}$ and the \textbf{Result Evaluator} $P_\theta^{eval}$, SeRTS iteratively performs \textit{Selection}, \textit{Expansion}, \textit{Evaluation}, and \textit{Backpropagation} until it finds a good enough query result or a computational limit is reached. The full pseudocode is provided in Appendix~\ref{sec:tsr_impl}.

\paragraph{Selection.} In the first operation, the algorithm selects the most promising leaf node for expansion, starting from the root node (initial state $s_0$). The Upper Confidence Bounds (UCB) is used to balance exploration and exploitation during node selection:
\begin{equation}
\begin{aligned}
    UCB(s)= \frac{R(s)}{N(s)} + C * \sqrt{2 * \ln \frac{N(p)}{N(s)}}
\end{aligned}
\end{equation}
, where $N(s)$ is the number of visits to a node $s$, $R(s)$ is the reward assigned to $s$, and $p$ is the parent node of $s$. At each level, the child node with the highest UCB value is selected. The $C$ parameter balances exploration and exploitation.

\paragraph{Expansion.} The selected node is expanded by sampling $n$ query proposals from $P_\theta^{query}$. The expansion is performed in a self-reflection manner to deal with the vast search space. All feedback $O^{sib}_f$ from $P_\theta^{eval}$, along with all the history of query proposals $O^{sib}_q$ from sibling nodes, as well as retrieved documents $O^{anc}_d$ and query proposals $O^{anc}_q$ from all the ancestor nodes, are fed into $P_\theta^{query}$ as reflection material as shown in Table~\ref{tab:query_prompt}. This practice ensures diverse reasoning by expanding different and potentially better branches after each sampling. 

\paragraph{Evaluation.} A scalar value is assigned to each new child node using feedback give by $P_\theta^{eval}$, quantifying the agent's progress in task completion and steering the search algorithm towards promising regions. Following previous work Self-rewarding Language Models~\cite{yuan2024self}, we assess the currently retrieved document set $D \cup O^{anc}_d$ with LLM-as-a-Judge prompting with a 5-point rubric as shown in Table~\ref{tab:eval_prompt}.

\paragraph{Backpropagation.} The average value and visit counts of the newly generated node $s'$ and its ancestors are updated using the scaled reward $R(s')$ from the Evaluation step.

By incorporating a self-rewarding paradigm and expanding strategy, SeRTS improves search efficiency and enhances the quality of retrieved documents. Details and pseudo-codes for SeRTS are illustrated in Appendix~\ref{sec:tsr_impl}.

\subsection{Proximal Policy Optimization of LLMs with Feedback}
To further improve the reasoning abilities of the Query Proposer and the Result Evaluator, we utilize Proximal Policy Optimization (PPO) to fine-tune the two LLMs using the trajectories collected by SeRTS as feedback~\cite{schulman2017proximal}. Additionally, through this process, we could validate the effectiveness of the SeRTS reasoning process by examining the trajectories it produces.

PPO aims to optimize the policy of an LLM agent by maximizing the expected cumulative reward while ensuring that the LLMs are updated conservatively in case of performance degradation. The objective function of PPO can be formulated as follows:
\vspace{-1mm}
\begin{equation}
    \begin{aligned}
    &\resizebox{1.0 \linewidth}{!}{$L^{ppo}(\theta)= \hat{\mathbb{E}}\left[\min \left(r(\theta) \hat{A}_t, \operatorname{clip}(r(\theta), 1-\epsilon, 1+\epsilon) \hat{A}_t\right)\right]$},\\
    &\mathrm{where}\ r_{\theta} = \frac{\pi_\theta\left(A_{1..i} \mid O_{1..i}\right)}{\pi_{\theta_{old}}\left(A_{1..i} \mid O_{1..i}\right)}
    \end{aligned}
\end{equation}
Here $\theta$ represents the parameters of the policy network, $r_{\theta}$ indicates the ratio of the probability of selecting an action under the new policy to that under the old one, and $A_{1..i}$ and $O_{1..i}$ denotes sampled query proposals and previous observations. $\hat{A}_t$ is the advantage function estimating the relative reward of taking action $A_{1..i}$ in the state of $O_{1..i}$ compared to the average value of the state, and $\epsilon$ is the clipping range hyperparameter.

During training, we sample trajectories $\mathcal{T}_{SeRTS}={(A_{1..i}, O_{1..i}, R_{1..i})}$ from the SeRTS retrieval process and the generated answers to estimate the advantage function and update the policy network $\theta$. The clipping function $\operatorname{clip}$ in $L^{ppo}(\theta)$ ensures that the policy update is conservative, preventing drastic changes that could lead to instability or performance degradation.

\begin{table*}[!t]
    \centering
    \small
    \setlength{\tabcolsep}{3.2mm}{
    \begin{tabular}{lcccc}
    \toprule
         Retrieval Method  
         &  Precision & Recall & F1 score & Hit Rate \\
         \midrule
         \textsc{BM25}  & 72.48 & 47.16 & 45.87 & 83.00 \\
         \midrule 
         \midrule
         \textsc{GPT-3.5-turbo + Reflection~(sim 6)}  & {66.26} & {58.96} & {51.18} & {88.89} \\
         \textsc{GPT-3.5-turbo + SeRTS~(sim 6, max ucb)} & \textbf{74.31} & \textbf{59.09} & \textbf{55.06} & \textbf{92.22} \\
         \midrule 
         \textsc{LLama2-7b + Reflection~(sim 6)}  & \textbf{72.48} & 48.66 & {47.12} & \textbf{84.44} \\
         \textsc{LLama2-7b + SeRTS~(sim 6, max ucb)}  & {67.85} & \textbf{50.95} & \textbf{47.58} & {83.33} \\
         \midrule 
         \midrule
         \textsc{GPT-3.5-turbo + Reflection~(sim 12)}  & 71.36 & 58.84 & 52.72 & 91.11 \\
         \textsc{GPT-3.5-turbo + SeRTS~(sim 12, max ucb)} & \textbf{74.65} & \textbf{58.96} & \textbf{54.70} & \textbf{94.44} \\
         \midrule 
         \textsc{LLama2-7b + Reflection~(sim 12)}  & 68.04 & 51.82 & 48.79 & 83.33 \\
         \textsc{LLama2-7b + SeRTS~(sim 12, max ucb)}  & \textbf{70.62} & \textbf{52.51} & \textbf{49.39} & \textbf{85.55} \\
        \bottomrule

    \end{tabular}
    }
    \caption{Retrieval Results on BioASQ Task 11b batch 1 Test Set. The highest values are \textbf{bolded}. All metrics are averaged across three runs with three random seeds~\{42, 43, 44\}.}
    \label{tab:retrieval_res}
\end{table*}

By updating the language model using PPO with the feedback $\mathcal{T}_{SeRTS}$, we aim to let LMs know how to make good query proposals given current observations. This training process allows the LLMs to adapt to the specific requirements of medical knowledge retrieval, self-improving their ability of RAG in the bio-medical domain. The fine-tuning and trajectory details are listed in \ref{sec:ppo_detail} and \ref{sec:sample_ppo}.

\section{Experiments}
\subsection{Datasets}
The BioASQ-QA~\cite{krithara2023bioasq} dataset is a comprehensive resource designed to meet the authentic information needs of biomedical professionals. It comprises questions in English, accompanied by gold-enriched answers by biomedical experts and relevant supporting material. This dataset stands out from most existing biomedical QA datasets due to its realistic and challenging nature, reflecting the complexity of real-world biomedical inquiries. Most importantly, BioASQ provides ideal answers in the form of summaries rather than only exact answers. This unique feature makes the dataset particularly valuable for researchers working on multi-document summarization techniques in the biomedical domain.
Furthermore, the BioASQ-QA dataset is updated annually through an open-challenge format, making it highly unlikely for LLMs to have accessed the data during training. To evaluate our approach, we utilize the BioASQ 2023 Task 11b test set~\cite{nentidis2023overview}, which consists of question-answer pairs along with their corresponding retrieved documents.

\subsection{Baselines}
We compare our proposed SeRTS method with Self-Reflection baselines. For a fair comparison, we implement Self-Reflection by making minor modifications to the SeRTS prompt. Specifically, we remove the sibling node information ($O^{sib}_f$ and $O^{sib}_d$) from the query proposer agent $P_{\theta}^{query}$ prompt, as self-reflection does not involve sibling nodes during expansion. The evaluator agent $P_{\theta}^{eval}$ prompt remains unchanged. We measure the retrieval capability of all methods by indexing the ground truth documents. Appendix~\ref{sec:refl_impl} provides details and pseudo-code for the self-reflection process.

\subsection{Metrics}
We evaluate retriever performance using standard metrics: \textbf{Precision}, \textbf{Recall}, and \textbf{F1 score}, comparing retrieved document IDs with the ground truth. \textbf{Hit Rate} assesses robustness by measuring the ratio of questions with at least one ground-truth document retrieved.

To measure RAG quality, we use \textbf{ROUGE-2 (R-2)} and \textbf{ROUGE-SU4 (R-SU4)}. R-2 computes shared bigrams between predicted answers and ground truth, while R-SU4 calculates skip-bigram overlap with a skip distance of four. Higher ROUGE scores indicate greater content overlap and more informative answers, suggesting better retrieval methods.

\subsection{Main Results}
\subsubsection{Comparison of the Retrieval Capability}

Table~\ref{tab:retrieval_res} shows that \textsc{SeRTS~(max ucb)} consistently improves \textbf{Recall} and \textbf{Hit Rate} compared to \textsc{Reflection} for both \textsc{GPT-3.5-turbo} and \textsc{LLama2-7B-chat} models. This performance gap supports our DoT assumption that higher-quality feedback induces more accurate reasoning. The performance gap between \textsc{GPT-3.5-turbo} and \textsc{LLama2-7B-chat} is attributed to \textsc{GPT-3.5}'s larger parameter count (175B) and stronger reasoning capacity compared to \textsc{LLama2} (7B). In addition, increasing the maximal simulation number further boosts the performance of both \textsc{SeRTS} and \textsc{Reflection}. \textsc{SeRTS (sim 12, max ucb)} improves \textbf{Recall} and \textbf{Hit Rate} by 5.5 and 2.5 percentage points compared to the vanilla \textbf{BM25} retriever.

\begin{table*}[!ht]
    \centering
    \small
    \setlength{\tabcolsep}{3.2mm}{
    \begin{tabular}{lcccc}
    \toprule
         Retrieval Method  
         &  R-2 Recall & R-2 F1 & R-SU4 Recall & R-SU4 F1 \\
         \midrule
         \textsc{Ground Truth}  & 57.02 & 56.64 & 58.26 & 57.42 \\
         \textsc{BM25}  & 25.58 & 22.74 & 29.63 & 26.22 \\
         \midrule
         \midrule
         \textsc{GPT-3.5-turbo + Reflection}  & 26.92 & 24.08 & 30.86 &  27.41 \\
         \textsc{GPT-3.5-turbo + SeRTS~(max ucb)} & \textbf{27.25} & \textbf{24.35} & \textbf{31.25} & \textbf{27.77} \\
         \midrule 
         \textsc{LLama2-7b + Reflection}  & 25.73 & 22.86 & 29.70 & 26.41 \\
         \textsc{LLama2-7b + SeRTS~(max ucb)}  & \textbf{26.18} & \textbf{23.20} & \textbf{30.30} & \textbf{26.79} \\
         \midrule 
         \textsc{LLama2-7b + Reflection~(PPO)}  & 26.20 & 23.41 & 30.24 & 26.93 \\
         \textsc{LLama2-7b + SeRTS~(PPO)}  & \textbf{26.51} & \textbf{23.63} & \textbf{30.73} & \textbf{27.34} \\
        \bottomrule

    \end{tabular}
    }
    \caption{Answer Generation Results on BioASQ Task 11b batch 1 Test Set. The highest values are \textbf{bolded}. All metrics are averaged across three runs with three random seeds~\{42, 43, 44\}.}
    \label{tab:main_exp}
\end{table*}

\subsubsection{Effect on RAG}
Table~\ref{tab:main_exp} presents the RAG performance of \textsc{LLama2-7b-chat} and \textsc{GPT-3.5-turbo} using different retrieval methods on the bio-medical QA task. The large gap between \textbf{BM25} and \textbf{GROUND TRUTH} highlights the importance of document retrieval quality in generating final answers for the difficult medical QA task. Variants of \textbf{SeRTS} perform best among all methods on all metrics, with $\textsc{SeRTS~(max ucb)}$ achieving 0.5 \textbf{R-2 F1} and 0.5 \textbf{R-SU4 F1} score improvements over \textsc{BM25} on BioASQ dataset. The improvement of \textsc{SeRTS~(max ucb)} over \textsc{Reflection} demonstrates the effectiveness of our proposed method.

Fine-tuning with the PPO objective further improves the performance of \textsc{Reflection} and \textsc{SeRTS}, with \textsc{SeRTS~(PPO)} acquiring 0.4 \textbf{R-2 F1} and 0.6 \textbf{R-SU4 F1} uplift over \textsc{SeRTS~(MAX UCB)} on \textsc{LLama2-7b-chat}. This suggests that PPO fine-tuning enables self-improvement and domain knowledge acquisition for bio-medical RAG. \textsc{SeRTS~(max ucb)} achieves 1.5 points \textbf{R-SU4 F1} increase over \textsc{BM25} using \textsc{GPT-3.5-turbo} model, while \textsc{LLama2-7b-chat + SeRTS~(PPO)} brings a total of 1.1 point improvement, narrowing the retrieval performance gap between the small and large language models.

The superiority of \textsc{SeRTS} method over \textsc{Reflection} is evident in both \textsc{PPO} and zero-shot variants when computation amounts are controlled to be similar (\textsc{sim 6} \& \textsc{sim 12}), showing the high scalability of \textsc{SeRTS} compared to \textsc{Reflection}. The detailed statistics of SeRTS's computational overhead can be found in Appendix~\ref{sec:overhead}.
\subsection{Ablation Study}

\subsubsection{Enhancement of PPO on SeRTS}
We examine the effect of PPO fine-tuning on RAG performance by gradually switching the original weight $\theta$ of the proposer $P_\theta^{query}$ and evaluator $P_\theta^{eval}$ to $\theta_p$ fine-tuned by PPO. Table~\ref{tab:weight_change} shows that switching from $P_\theta^{query}$ to $P_{\theta_p}^{query}$ increases \textbf{R-SU4 F1} from 26.41 to 26.91 in \textsc{Reflection} and from 26.79 to 26.89 in \textsc{SeRTS}. Further switching from $P_\theta^{eval}$ to $P_{\theta_p}^{eval}$ with $P_{\theta_p}^{query}$ as the query proposer continues to improve performance on all metrics. This suggests that $P_{\theta_p}^{eval}$ has learned to better evaluate retrieved documents and provide suggestions for improved queries, possibly due to the fact that the LM learns from feedback observations $F$ and rewards $R$ in trajectories $\mathcal{T}_{SeRTS}$ about what feedback and suggestions can lead to better queries from the proposer $P_\theta^{query}$.

\begin{table*}[!ht]
    \centering
    \small
    \setlength{\tabcolsep}{3.2mm}{
    \begin{tabular}{lcccc}
    \toprule
         Evaluating Agent  
         &  R-2 Recall & R-2 F1 & R-SU4 Recall & R-SU4 F1 \\
         \midrule
         \textsc{+ Reflection} & 25.73 & 22.86 & 29.70 & 26.41\\
         \midrule
         \textsc{+ Reflection~(PPO)}  & \multirow{2}{*}{26.15} & \multirow{2}{*}{23.36} & \multirow{2}{*}{30.11} & \multirow{2}{*}{26.83} \\
         \textsc{w. original evaluator} & ~ & ~ & ~ & ~\\
         \textsc{+ Reflection~(PPO)}  & \multirow{2}{*}{\textbf{26.20}} & \multirow{2}{*}{\textbf{23.41}} & \multirow{2}{*}{\textbf{30.24}} & \multirow{2}{*}{\textbf{26.93}} \\
         \textsc{w. PPO evaluator} & ~ & ~ & ~ & ~\\
         \midrule\midrule
         \textsc{+ SeRTS} & 26.18 & 23.20 & 30.30 & 26.79\\
         \midrule
         \textsc{+ SeRTS~(PPO)}  & \multirow{2}{*}{25.90} & \multirow{2}{*}{23.42} & \multirow{2}{*}{30.34} & \multirow{2}{*}{26.89} \\
         \textsc{w. original evaluator} & ~ & ~ & ~ & ~\\
         \textsc{+ SeRTS~(PPO)}  & \multirow{2}{*}{\textbf{26.51}} & \multirow{2}{*}{\textbf{23.63}} & \multirow{2}{*}{\textbf{30.73}} & \multirow{2}{*}{\textbf{27.34}} \\
         \textsc{w. PPO evaluator} & ~ & ~ & ~ & ~\\
         
        \bottomrule

    \end{tabular}
    }
    \caption{Effect on Answer Generation Performance of PPO Finetuning on \textsc{Llama2-7b-chat}. The highest values are \textbf{bolded}. All metrics are averaged across three runs with three random seeds~\{42, 43, 44\}.}
    \label{tab:weight_change}
\end{table*}

\subsubsection{Difference on Selection Policy}

\begin{table*}[!ht]
    \centering
    \small
    \setlength{\tabcolsep}{3mm}{
    \begin{tabular}{lcccc}
    \toprule
         Retrieval Method  
         &  R-2 Recall & R-2 F1 & R-SU4 Recall & R-SU4 F1 \\
         \midrule
         \textsc{BM25} & 25.58 & 22.74 & 29.63& 26.22 \\
         \textsc{Refelction}  & 26.92 & 24.08 & 30.86 & 27.41 \\
         \midrule\midrule
         \textsc{SeRTS~(random)}  & 25.74 & 22.87 & 29.73 & 26.29 \\
         \textsc{SeRTS~(max ucb, $C=0.0$)}  & 25.73 & 22.87 & 29.70 & 26.27 \\
         \textsc{SeRTS~(max ucb, $C=0.1$)}  & 
         \textbf{27.25} & \textbf{24.35} & \textbf{31.25} & \textbf{27.77} \\
    \bottomrule
    \end{tabular}
    }
    \caption{Effect on Answer Generation Performance of Selection Policy in \textsc{SeRTS}. The highest values are \textbf{bolded}. All metrics are averaged across three runs with three random seeds~\{42, 43, 44\}.}
    \label{tab:param_select}
\end{table*}

    
    
Table~\ref{tab:param_select} examines how different selection policies in \textsc{SeRTS} affect retrieval performance. Here we evaluate three policies: 1)~\textsc{random}, which randomly explores nodes without exploiting assigned scores $R$; 2)~\textsc{max ucb~($C=0.0$)}, which only selects nodes with the highest $R$ without exploring other options; and 3)~\textsc{max ucb~($C=0.1$)}, which balances exploitation and exploration. The results empirically prove that striking a balance between exploitation and exploration is necessary for \textbf{SeRTS} to achieve ideal performance, while pure exploitation or exploration causes performance degradation.

\section{Conclusion}
We propose Self-Rewarding Tree Search~(SeRTS), a plug-and-play retrieval method that combines LLM reasoning with conventional retrievers to address limitations in biomedical retrieval-augmented approaches. SeRTS introduces a reasoning process based on Monte Carlo Tree Search (MCTS) and a self-rewarding paradigm, boosting zero-shot performance in retrieving high-quality, informative results for Retrieval Augmented Generation (RAG). We further enhance retrieval by fine-tuning LLMs with PPO objectives using SeRTS retrieval trajectories as feedback.

Experiments using the BioASQ-QA dataset with GPT-3.5-Turbo and LLama2-7b show that SeRTS significantly improves BM25 retriever performance, surpassing the self-reflection baseline in efficiency and scalability. SeRTS (PPO) effectively adapts LLMs to document retrieval, enhancing their ability to find highly relevant documents for RAG when answering medical queries.
Our work presents an effective method for accurate, comprehensive medical knowledge retrieval using LLMs. While SeRTS outperforms Self-Reflection, both require numerous simulations, limiting widespread application. Future research should focus on developing more efficient reasoning strategies and leveraging the self-improvement capabilities of LLMs to further enhance retrieval efficiency and usability.

\section*{Limitations}
PPO fine-tuning effectively improves SeRTS's performance but is prone to Catastrophic Forgetting, degrading LLMs' instruction-following capability. Table~\ref{tab:error_case} shows error cases where $P_\phi^{query}$ fails to follow the designed instruction. Although manual templates can temporarily address these issues in our current experiments, the noisy trajectories impede further iteration and self-improvement using SeRTS sampling and PPO fine-tuning when fed back into the loop. As a result, we restrict our experiments to a single iteration of sampling and PPO fine-tuning. Future research should prioritize mitigating Catastrophic Forgetting and investigate the potential of multiple PPO fine-tuning iterations for further LLM self-improvement.
\begin{table}
\small
    \centering
    \colorbox{orange!8}{
    \begin{tabular}{@{}p{7.2cm}}
    Based on the retrieved information, I would suggest the following query to search for more information:\\
    \textcolor{green!50!black}{<query> Does losartan slow down brain atrophy in Alzheimer's disease? A systematic review and meta-analysis. </query>}\\
    \\
    Based on the information retrieved, I would suggest the following query to further search for relevant information:\\
    \textcolor{red!70!black}{"Does losartan slow down brain atrophy in Alzheimer's disease?"}\\
    \\
    Based on the retrieved information, I would suggest the following query to retrieve more abstracts:\\
    \textcolor{red!70!black}{<query> Does losartan slow down brain atrophy in Alzheimer's disease? A systematic review and meta-analysis. Corresponding Feedback: Total Score: 40/50}\\
    \\
    Based on the retrieved information, I would suggest the following query to further search for relevant information:\\
    \textcolor{red!70!black}{"Losartan and brain atrophy in Alzheimer's disease" Query Here}\\
    \\
    Based on the retrieved information, I would suggest the following query to further search for additional information on dermatillomania:\\
    \textcolor{red!70!black}{"<query> Causes of dermatillomania Query Here >"}\\
    \end{tabular}}
    \caption{Examples of $P_\phi^{query}$ generating query proposals. Cases where $P_\phi^{query}$ successfully follows the instruction are marked in \textcolor{red!70!black}{red}, while cases where $P_\phi^{query}$ fails to follow the instruction are marked in \textcolor{green!50!black}{green}.}
    \label{tab:error_case}
\end{table}

\section*{Acknowledgement}
The work described in this paper was supported by the Chinese University of Hong Kong (CUHK; award numbers 4937025, 4937026, 5501517 and 5501329 to Y.L.) and the IdeaBooster Fund (IDBF23ENG05 and IDBF24ENG06 to Y.L.) and partially supported by grants from the Research Grants Council of the Hong Kong Special Administrative Region (Hong Kong SAR), China (project no. CUHK 14222922, RGC GRF 2151185 to I.K., and CUHK 24204023 to Y.L.) and a grant from the Innovation and Technology Commission of the Hong Kong SAR, China (project no. GHP/065/21SZ to Y.L.). This research was also funded by the Research Matching Grant Scheme at CUHK (award numbers 8601603 and 8601663 to Y.L.). 
\bibliography{anthology,emnlp2023-latex/custom_mcts_rebibed}
\bibliographystyle{acl_natbib}

\appendix
\section{Five-Point Rubrics of Evaluator}
\label{sec:rubric}
Details of evaluation rubrics are illustrated in Table~\ref{tab:rubrics}.
\begin{table}[H]
\small
    \centering
    \colorbox{orange!8}{
    \begin{tabular}{@{}p{7.2cm}}
    - Add 1 point if the knowledge is relevant and provides some information related to the user's inquiry, even if it is incomplete or contains some irrelevant content.\\
    - Add another point if the knowledge addresses a substantial portion of the user’s question, but does not completely resolve the query or provide a direct answer.\\
    - Award a third point if the response answers the basic elements of the user’s question in a useful way, regardless of whether it seems to have been written by an AI Assistant or if it has elements typically found in blogs or search results.\\
    - Grant a fourth point if the knowledge addresses the user’s question directly and comprehensively, even if there is slight room for improvement in clarity, conciseness, or focus.\\
    - Bestow a fifth point for knowledge that is impeccably tailored to the user’s question, without extraneous information, reflecting expert knowledge, and can be used to produce a high-quality, engaging, and insightful answer.
    \end{tabular}}
    \caption{Details of \textbf{\{Five-point Rubrics\}}}
    \label{tab:rubrics}
\end{table}
\section{Statistics of Computational Overhead}
\label{sec:overhead}
To show how much overhead is introduced, we measured the mean number of tokens per question generated by the \textsc{LLAMA2-7B-CHAT} model for both our method (SeRTS) and the Self-Reflection baseline during the retrieval and reasoning process. The results are Table~\ref{tab:token_num}.

\begin{table}[!ht]
    \centering
    \setlength{\tabcolsep}{2mm}{
    \begin{tabular}{lc}
    \toprule
         Retrieval Method  
         &  Avg. \#Token\\
         \midrule
         \textsc{Self-Reflection~(sim 6)} & 4691.0  \\
         \textsc{SeRTS~(sim 6)}  & 4758.3\\
         \midrule
         \textsc{Self-Reflection~(sim 12)}  & 13,330.3\\
         \textsc{SeRTS~(sim 12)}  & 9,309.4\\ 
         \textsc{SeRTS~(sim 12, PPO)} & 5,175.9\\ 
    \bottomrule

    \end{tabular}
    }
    \caption{Average number of tokens per question generated by \textsc{LLama2-7b-chat} during the experiment in Table~\ref{tab:main_exp}.}
    \label{tab:token_num}
\end{table}
\textsc{SeRTS} significantly reduces the number of generated tokens compared to \textsc{Self-Reflection}. With a simulation round of 12, \textsc{SeRTS} generates 30.16\% fewer tokens than \textsc{Self-Reflection}. When incorporating PPO into \textsc{SeRTS}, the token reduction achieves 57.12\%. 

\section{Experiments on MedMCQA dataset}
We conducted experiments on the MedMCQA~\cite{pmlr-v174-pal22a} dataset to demonstrate the robustness of \textsc{SeRTS}. Using \textsc{GPT-3.5-Turbo} as the answer generator, we sample 100 questions from the development set and compare four configurations: \textsc{GPT-3.5} without retrieval, \textsc{GPT-3.5} with BM25 retriever, \textsc{Self-Reflection~(Sim 12)}, and \textsc{SeRTS~(Sim 12)}.\textsc{SeRTS} achieves an accuracy of 0.57, outperforming the no-retrieval baseline by 7.55\% and \textsc{Self-Reflection} by 3.64\%, demonstrating its effective generalization to other biomedical QA datasets.
\begin{table}[!ht]
    \centering
    \setlength{\tabcolsep}{2mm}{
    \begin{tabular}{lc}
    \toprule
        Method  
         &  Accuracy\\
         \midrule
         \textsc{GPT-3.5 wo. retrieval} & 0.53  \\
         \textsc{GPT-3.5 w. retrieval}  & 0.55\\
         \textsc{GPT-3.5 w. Reflection}  & 0.55\\
         \textsc{GPT-3.5 w. SeRTS}  & \textbf{0.57}\\ 
    \bottomrule

    \end{tabular}
    }
    \caption{Performance of multi-choice QA on MedMCQA dataset.}
    \label{tab:medmcqa_exp}
\end{table}

\section{Implementation Details}

\subsection{RAG Pipeline}
It searches for relevant text from the document test set using the questions as queries and supplies these texts as references to the language model. The language model we used is the \textsc{LLama2-7B-chat}~\cite{touvron2023llama2}, which has been fine-tuned on the BioASQ corpus using LoRA~\cite{hu2021lora}. This corpus comprises 4,719 question-answer pairs from the BioASQ task 11b training set. Importantly, the documents in the training set are distinct from those in the test set, ensuring no overlap.
 The BM25 retrieval function can be expressed as follows:
\begin{equation}
    \begin{aligned}
    &\text{score}(q, d) = \sum_{i=1}^{n} IDF(q_i)\\
    &\cdot \frac{f(q_i, d) \cdot (k_1 + 1)}{f(q_i, d) + k_1 \cdot \left(1 - b + b \cdot \frac{|d|}{\text{avgdl}}\right)}
    \end{aligned}
\end{equation}
, where
 \( q \) is the query, and \( d \) is the document.
 \( q_i \) is the \( i \)-th term in the query.
 \( f(q_i, d) \) is the term frequency of \( q_i \) in the document \( d \).
 \( |d| \) is the length of the document \( d \) in terms of the number of words.
 \( \text{avgdl} \) is the average document length in the collection.
 \( k_1 \) and \( b \) are free parameters, usually chosen, for example, \( k_1 = 1.2 \) and \( b = 0.75 \).
 \( IDF(q_i) \) is the inverse document frequency weight of the term \( q_i \), calculated as:
\begin{equation}
IDF(q_i) = \log \left( \frac{N - n(q_i) + 0.5}{n(q_i) + 0.5} + 1 \right)  
\end{equation}
, where
 \( N \) is the total number of documents in the collection.
 \( n(q_i) \) is the number of documents containing the term \( q_i \).

To minimize the influence of the prompt, we employed a very simple prompt to instruct the language model to generate appropriate answers based on the provided text. The structure of this prompt is illustrated as follows.
\begin{table}[!ht]
\small
    \centering
    \colorbox{orange!8}{
    \begin{tabular}{@{}p{7.2cm}}
    You are an expert in the field of biomedical science.\\
    \\
    Below is an instruction that describes a task, paired with an input that provides further context. Write a response that appropriately completes the request.\\
    \\
    \#\#\# Instruction:
    \textbf{\{question\}}\\
    \\
    \#\#\# Input:
    \textbf{\{context\}}\\
    \\
    \#\#\# Response: \\
    \end{tabular}}
    \caption{Prompt for Question Answering.}
    \label{tab:question_prompt}
\end{table}
\subsection{Self-Rewarding Tree Search (SeRTS)}
We use the Llama2-7b-chat model published in the Huggingface~\footnote{\url{https://huggingface.co/meta-llama/Llama-2-7b-chat-hf}} for experiments in this work. The Llama2-7b-chat model is loaded in 4 bits, and it uses flash attention 2 as attention implementation for efficiency. For sequence generation, we keep the same configuration of setting the temperature to 0.7 and maximal length to 4096 tokens. For all experiments, we report the averaged metrics from three runs with the random seed fixed to \{42,43,44\}. For the SeRTS process illustrated in Algorithm~\ref{algo:tsr}, we set the maximal simulation number $Sim$ to 12, maximal branch number $b$ to 3, and maximal depth $D$ to 3.
\label{sec:tsr_impl}
\begin{algorithm*}
\caption{The Self-Rewarding Tree Search~(SeRTS) Algorithm.}
\label{algo:tsr}
\begin{algorithmic}[1]
\State {\bfseries Input:}  question $Q$, query proposer $P_\theta^{query}$, evaluator $P_\theta^{eval}$, max\_simulation\_number $Sim$, max\_branch\_number $b$, max\_depth $Depth$, BM25\_retriever $\mathcal{R}_{bm25}$.
\State $T_Q \gets \text{Initialize\_tree}(Q)$
\For{$i$ in range($Sim$)}
    \State $C \gets \text{root}(T_q)$
    \State \textbf{\textit{---------------------Selection---------------------}}
    \While{$C$ is not leaf node do}
        \State $C \gets \arg\max_{s} UCB(s)= \frac{R(s)}{N(s)} + C * \sqrt{2 * \ln \frac{N(p)}{N(s)}}$ \Comment{Select child node based on UCB}
        \If{$C$'s depth >= $Depth$} \Comment{Break if $C$ reaches maximal depth}
        \State \textbf{break}
        \EndIf
        \\
        \If{$C$ has less than $b$ children nodes} \Comment{Select $C$ if it is not fully expanded}
        \State \textbf{break}
        \EndIf
    \EndWhile
    \If{$C$'s depth >= $Depth$} \Comment{Continue to select other nodes if $C$ reaches maximal depth}
        \State \textbf{continue}
    \EndIf
    \\
    \State \textbf{\textit{------------Expansion \& Evaluation------------}}
    
    \State $O^{anc}_q \gets \bigcup_{k \in \text{ancestors}(C)}A_k: \text{query proposals from all $C$'s ancestor nodes}$
    \State $O^{anc}_d \gets \bigcup_{k \in \text{ancestors}(C)}D_k: \text{retrieved documents from all $C$'s ancestor nodes}$
    \State $O^{sib}_q\ \gets \bigcup_{k \in \text{siblings}(C)}A_k: \text{query proposals from all $C$'s sibling nodes}$
    \State $O^{sib}_f\ \gets \bigcup_{k \in \text{siblings}(C)}F_k: \text{feedback from all $C$'s sibling nodes}$
    \State $O \gets (O^{anc}_q, O^{anc}_d, O^{sib}_q, O^{sib}_f)$
    \State $A \gets P_\theta^{query}(O, Q)$ \Comment{Make new query proposal $A$ based on $O$ and $Q$}
    \State $D \gets R_{bm25}(A)$ \Comment{Invoke $\mathcal{R}_{bm25}$ to retrieve documents according to $A$}
    \State $R, F \gets P_\theta^{eval}(D \cup O_d^{anc}, Q)$ \Comment{Get reward score $R$ and descriptional feedback $O_f$ for currently retrieved documents $D \cup O_d^{anc}$ and $Q$}
    \If{$R$ == 5} \Comment{Stop early if $D \cup O_d^{anc}$ receive highest score in 5-point rubrics}
        \State \textbf{break}
    \EndIf
    \\
    \State $C' = \text{create\_new\_child}(Q, A, R, O)$ \Comment{create a new child node for $C$}
    \State Add $C'$ to children nodes of $C$
    \State \textbf{\textit{------------------Backpropagation------------------}}
    \State $\text{back\_propagate}(C)$ \Comment{Update value of parent nodes using new score from $C'$}
\EndFor
\State $C = \text{get\_best\_node}(T_Q)$ \Comment{Fetch the node with the highest $R$ in $T_Q$}
\State \Return $D \cup O_d^{anc}$ from $C$ \Comment{Return all retrieved documents in $C$}
\end{algorithmic}
\end{algorithm*}
\subsection{Self-Reflection}
Compared with prompt for $P_\theta^{query}$ in Table~\ref{tab:query_prompt}, prompt for $P_\theta^{reflect}$ in Table~\ref{tab:refl_prompt} removes all feedback from sibling nodes. Algorithm~\ref{algo:reflect} shows a detailed implementation of the self-reflection process, node selection, and backpropagation operation is removed compared to SeRTS. All other configurations of the self-reflection process remain unchanged.
\begin{table}[!ht]
\small
    \centering
    \colorbox{orange!8}{
    \begin{tabular}{@{}p{7.2cm}}
    You are a medical expert, and you are tasked with searching for as much useful knowledge from a database of medical paper abstracts as possible to address the user’s question. The question is stated as follows: \textbf{\{$Q$\}}\\
    \\
    You are given the BM25 retriever as the search tool. According to previous query proposals and proposals' corresponding feedback and suggestions, give one single better query for the retriever to further search for extra information needed for answering the given question.\\
    \\
    Information already known is as follows:<retrieved\_info>\\
    \ \ \ \ Query and its reasoning for Abstracts: \textbf{\{$O^{anc}_q$\}}\\
    \ \ \ \ Retrieved Abstracts: \textbf{\{$O^{anc}_d$\}}\\
    
    </retrieved\_info>\\ 
    \\
    After examining the information above, justify your decision, in up to 100 words.\\
    \\
    Hints: \\
    1. Use simple keywords as queries if no information is retrieved.\\
    2. You can take one step at a time, subdivide the original question, and search for information on the sub-questions.\\
    \\
    Conclude with the query for the paper abstracts using the format: "<query> Query Here </query>".
    \end{tabular}}
    \caption{Prompt for Self-Reflect Query Proposer $P^{reflect}_{\theta}$.}
    \label{tab:refl_prompt}
\end{table}

\label{sec:refl_impl}
\begin{algorithm*}
\caption{The Self-Reflection Algorithm.}
\label{algo:reflect}
\begin{algorithmic}[1]
\State {\bfseries Input:}  question $Q$, self-reflect query proposer $P_\theta^{reflect}$, evaluator $P_\theta^{eval}$, max\_simluation number $Sim$, max\_branch\_number $b$, max\_depth $Depth$, BM25\_retriever $\mathcal{R}_{bm25}$.
\State $T_Q \gets \text{Initialize\_tree}(Q)$
\For{$i$ in range($Sim$)}
    \State $C \gets \text{root}(T_q)$
    \State \textbf{\textit{------------Expansion \& Evaluation------------}}
    
    \State $O^{anc}_q \gets \bigcup_{k \in \text{ancestors}(C)}A_k: \text{query proposals from all $C$'s ancestor nodes}$
    \State $O^{anc}_d \gets \bigcup_{k \in \text{ancestors}(C)}D_k: \text{retrieved documents from all $C$'s ancestor nodes}$
    \State $O \gets (O^{anc}_q, O^{anc}_d)$
    \State $A \gets P_\theta^{reflect}(O, Q)$ \Comment{Make new query proposal $A$ based on $O$ and $Q$}
    \State $D \gets \mathcal{R}_{bm25}(A)$\Comment{Invoke $R_{bm25}$ to retrieve documents according to $A$}
    \State $R, F \gets P_\theta^{eval}(D \cup O_d^{anc}, Q)$\Comment{Get reward score $R$ and descriptional feedback $F$ for currently retrieved documents $D \cup O_d^{anc}$ and $Q$}
    \If{$R$ == 5} \Comment{Stop early if $D \cup O_d^{anc}$ receive highest score in 5-point rubrics}
        \State \textbf{break}
    \EndIf
    \\
    \State $C' = \text{create\_new\_child}(Q, A, R, O)$ \Comment{create a new child node for $C$}
    \State Add $C'$ to children nodes of $C$
\EndFor
\State $C = \text{get\_best\_node}(T_Q)$ \Comment{Fetch the node with the highest $R$ in $T_Q$}
\State \Return $D \cup O_d^{anc}$ from $C$ \Comment{Return all retrieved documents in $C$}
\end{algorithmic}
\end{algorithm*}
\subsection{LoRA PPO Finetuning}
\label{sec:ppo_detail}

For PPO fine-tuning, we utilize the LoRA and PPO implementations from the peft and trl Python packages, following the official training script\footnote{\url{https://github.com/huggingface/trl/blob/main/examples/research_projects/stack_llama/scripts/rl_training.py}}. The fine-tuning process is completed on a single Nvidia A6000 GPU within 2 hours. The lora\_target\_modules is set to \textit{"["q\_proj", "v\_proj"]"}, with lora\_r as 16, lora\_alpha as 32, and lora\_dropout as 0.05. The micro\_batch\_size is 1, batch\_size is 8, and num\_epochs is 1. For optimization, we set the learning\_rate to 1.41e-5, the reward baseline to 3.75, and the random seed to 0.

\subsection{Training Sample of PPO Trajectory}
\label{sec:sample_ppo}
Table~\ref{tab:traj_demo} demonstrates a detailed sample trajectory from $\mathcal{T}_{SeRTS}$ used during PPO training.
\begin{table}[!ht]
\small
    \centering
    \colorbox{orange!8}{
    \begin{tabular}{@{}p{15.5cm}}
    \toprule
    \textbf{Observation $\mathcal{O}$:}\\
    \midrule
    You are a medical expert, and you are tasked with searching for as much useful knowledge from a database of medical paper abstracts as possible to address the user’s question. The question is stated as follows: \uline{What disease can be treated with Glofitamab?}\\\\
    You are given the BM25 retriever as the search tool. According to previous query proposals and proposals' corresponding feedback and suggestions, give one single better query for the retriver to further search extra information needed for answering the given question.\\\\
    \textbf{<query\_proposal>}\\
    \hangindent=2em 
    \hangafter=0 
    1.  What are the efficacy and safety of Glofitamab in treating B-cell lymphoma in patients who have failed traditional chemotherapy and CAR T therapy? \uline{Corresponding Feedback:} Total Score: 3. The response provides valuable information related to the disease that can be treated with Glofitamab. ... To improve the query and obtain more relevant results, the following suggestions could be made: * Use more specific keywords: Instead of using the generic term "disease," consider using more specific terms related to the types of lymphoma that can be treated with Glofitamab, such as "DLBCL" or "MCL." ...
    By incorporating these search terms into the query, the user may obtain more targeted and relevant results related to the use of Glofitamab in treating lymphoma.\\
\textbf{</query\_proposal>}.\\\\
 Information already known is as follows: \textbf{<retrieved\_info>}\\
 \quad Query and its reasoning for Abstracts: \textbf{<start>}\\
\hangindent=2em 
\hangafter=0 
1. What disease can be treated with Glofitamab?\\\\
\hangindent=2em 
\hangafter=0 
2. Based on the retrieved information, I would suggest the following query to further search for relevant information: "What are the efficacy and safety of Glofitamab in treating B-cell lymphoma in patients who have failed traditional chemotherapy and CAR T therapy?" ...
By answering this query, we can gain a better understanding of the potential benefits and risks of Glofitamab in treating B-cell lymphoma in patients who have failed previous treatments, and inform future clinical trials and treatment decisions.
\textbf{<query>} What are the efficacy and safety of Glofitamab in treating B-cell lymphoma in patients who have failed traditional chemotherapy and CAR T therapy? \textbf{</query>}\\
\quad\textbf{<end>}
\\\\
\quad Retrieved Abstracts: \textbf{<start>}\\
\hangindent=2em 
\hangafter=0 
1. Chimeric antigen receptor T-cells (CAR T) treatment has become a standard option for patients with diffuse large B-cell lymphomas (DLBCL), which are refractory or relapse after two prior lines of therapy...\\\\
\hangindent=2em 
\hangafter=0 
2. Mantle cell lymphoma (MCL) is a rare type of B-cell Non-Hodgkin lymphoma (NHL) affecting predominantly male patients. While complete remissions following first-line treatment are frequent, most patients ultimately relapse, with a usually aggressive further disease course...\\\\
\hangindent=2em 
\hangafter=0 
3. We present a novel approach for first-in-human (FIH) dose selection of the CD20xCD3 bispecific antibody, glofitamab, based on pharmacokinetic/pharmacodynamic (PKPD) assessment in cynomolgus monkeys to select a high, safe starting dose, with cytokine release (CR) as the PD endpoint...\\
\quad\textbf{<end>}\\
\textbf{</retrieved\_info>}\\
\\
 After examining the information above, justify your decision, in up to 100 words.\\\\
 Hints:\\
1. Use simple keywords as queries if no information is retrieved.\\
2. You can take one step at a time and subdivide the original question and search information of the subquestions.\\\\
Conclude with the query for the paper abstracts using the format: "\textbf{<query>} Query Here \textbf{</query>}"\\
    \toprule
    \textbf{Action $\mathcal{A}$}:\\
    \midrule
    Based on the retrieved information, the best query to further search for information on the disease that can be treated with Glofitamab is: "What are the different types of B-cell lymphoma that can be treated with Glofitamab?\\\\This query is more specific than the original question and can help retrieve more relevant information on the potential uses of Glofitamab in different types of B-cell lymphoma. The retrieved information mentions the use of Glofitamab in mantle cell lymphoma, but does not provide information on other types of B-cell lymphoma that it may be effective against. By searching for information on the different types of B-cell lymphoma that can be treated with Glofitamab, we can gain a better understanding of its potential therapeutic applications.\\\\Here is the query for the paper abstracts:\\"\textbf{<query>} What are the different types of B-cell lymphoma that can be treated with Glofitamab?\textbf{</query>}".\\
    \toprule
    \textbf{Reward $\mathcal{R}$}:\\
    \midrule
    3\\
    \end{tabular}
}
    
    \begin{tabular}{@{}p{16cm}}
    \caption{Sample trajectory $(\mathcal{O}, \mathcal{A}, \mathcal{R})$ from $\mathcal{T}_{SeRTS}$.}
    \label{tab:traj_demo}
    \end{tabular}
\end{table}

\end{document}